\documentclass[summary]{ursi}
\usepackage{booktabs}
\usepackage{multirow}
\usepackage{placeins}
\usepackage[style=ieee, natbib=true, backend=biber]{biblatex}
\title{Polarisation-Inclusive Spiking Neural Networks for Real-Time RFI Detection in Modern Radio Telescopes}

\author{Nicholas J. Pritchard*\affref{ref1}, Andreas Wicenec\affref{ref1}, Richard Dodson\affref{ref1} and Mohammed Bennamoun\affref{ref2}}

\affiliation{%
  \aff{ref1}{International Centre for Radio Astronomy Research, Perth, Australia}
  \aff{ref2}{The School of Physics, Mathematics and Computing, The University of Western Australia, Perth, Australia}
}

\begin{document}

\maketitle

\begin{abstract}
Radio Frequency Interference (RFI) is a known growing challenge for radio astronomy, intensified by increasing observatory sensitivity and prevalence of orbital RFI sources.
Spiking Neural Networks (SNNs) offer a promising solution for real-time RFI detection by exploiting the time-varying nature of radio observation and neuron dynamics together.
This work explores the inclusion of polarisation information in SNN-based RFI detection, using simulated data from the Hydrogen Epoch of Reionisation Array (HERA) instrument and provides power usage estimates for deploying SNN-based RFI detection on existing neuromorphic hardware.
Preliminary results demonstrate state-of-the-art detection accuracy and highlight possible extensive energy-efficiency gains.
\end{abstract}

\section{Introduction}
RFI detection for contemporary and future radio telescopes is becoming a major challenge.
With greater sensitivity and data volumes, observatories must balance rising data-processing requirements while addressing increasingly pervasive RFI sources and subject to operational constraints.
Spiking Neural Networks (SNNs) are a compelling approach towards real-time RFI detection due to their inherently time-varying nature and extreme energy efficiency when deployed on neuromorphic hardware \cite{davies_loihi_2018, bos_sub-mw_2022}.
However, unlike Artificial Neural Networks (ANNs), which communicate through weighted continuous and instantaneous values, SNNs communicate through strictly binary but asynchronous `spikes'.
Encoding and decoding information for use in SNNs is challenging, but if successful allows th=e network to exploit continuous time dynamics to process spatio-temporal information efficiently.
Building on recent advances in SNN-based RFI detection \cite{pritchard_rfi_2024, pritchard_supervised_2024, pritchard_spiking_2024}, this work investigates including polarisation information and further presents estimates of power-consumption when deployed on neuromorphic hardware.
Using a simulated dataset derived from the Hydrogen Epoch of Reionisation Array (HERA) simulator, we demonstrate state-of-the-art detection performance and compelling operational estimates using SynSense's Xylo hardware as a model deployment platform.

\section{RFI Detection with SNNs}
We utilise a known supervised time-series segmentation approach for RFI detection \cite{pritchard_supervised_2024} modified to include multiple polarisation channels:
\begin{equation}\label{eq:formulation_polarization}
    \mathcal{L}_{sup} = 
    min_{\theta_n}(
        \Sigma_t^T
        \mathcal{H}(
            m_{\theta_n}(
                E(V(\upsilon, p, t, b))
            ), 
            F(G(\upsilon, p, t, b))
        )
    )
\end{equation}
$\theta_n$ are the parameters of some classifier $m$, $\mathcal{H}$ is a comparator function between the supervision masks and network output. 
$V$ is the original visibilities, $G$ is the ground-truth, $E$ and $F$ are spike-encoding and decoding functions.
Finally, $\upsilon$ is the visibility frequency, $p$ the polarisation, $t$ the time and $b$ baseline in the original visibility data.
This formalisation introduces an exposure time $e$ for each time-step in the visibilities and permits flexibility in how the visibility data are encoded into spikes for an SNN.

We consider latency encoding specifically, which has previously been shown to be an effective spike encoding for this task \cite{pritchard_supervised_2024} and two methods to include polarisation information, directly including all four polarisations as separate input channels (XX, XY, YX and YY) and a Degree of Polarisation (DoP) metric \cite{hashimoto_rfi_2024}.
Moreover, we investigate the effect of a divisive normalisation technique inspired from neuroscience research that has been shown to be helpful in this task \cite{pritchard_spiking_2024}.
Divisive normalisation is a pre-processing technique applied before the calculation of the DoP metric.
The HERA dataset has been well-described elsewhere \cite{mesarcik_learning_2022, pritchard_supervised_2024} and we prepare a similar dataset but retain all polarisation channels, which are openly available online\footnote{\url{https://doi.org/10.5281/zenodo.14676274}}.
In addition to training SNNs on the full spectrogram size, like prior works \cite{mesarcik_learning_2022, pritchard_rfi_2024, pritchard_spiking_2024, pritchard_supervised_2024} we divide the original spectrograms into smaller patches for more manageable training, and in this work specifically, to conceivably fit on contemporary neuromorphic hardware.

We employed the snnTorch framework for SNN training \cite{eshraghian_training_2023}, with data-driven parallelism managed by the Lightning library \cite{falcon_pytorch_2019}.
The spiking neuron model is a first-order Leaky integrate and Fire (LiF) neuron.
The Xylo platform is intended for edge-inference of SNNs, it supports 16 input channels, 1000 internal neurons and 16 output channels \cite{noauthor_xylo-audio_2022}.
We trained SNNs including full-polarisation information, DoP information, both with and without divisive-normalisation and then for full spectrograms, $16 \times 16$ spectrogram patches and smaller $15 \times 16$ patches where including DoP and $2 \times 16$ patches where including all polarisations.
The network is otherwise a multi-layer feed-forward network of spiking neurons.
Table \ref{tab:models} summarises the model input and output sizes for each configuration tested.
In all cases, the number of hidden neurons is 512 and the final layer contains either 16, 4 or 8 neurons in the case of no, full or DoP polarisation information respectively.

\begin{table}[!htb]
\centering
\caption{Model parameters for networks tested. `Channels in' refers to number of spectrographic channels (exclusive of polarisation).}
\label{tab:models}
\begin{tabular}{@{}cccc@{}}
\toprule
Model Type                        & Polarisation & \begin{tabular}[c]{@{}c@{}}Channels\\ In\end{tabular} & \begin{tabular}[c]{@{}c@{}}Channels\\ Out\end{tabular} \\ \midrule
\multirow{2}{*}{Patched}          & DoP          & 16                                                    & 16                                                     \\
                                  & Full         & 64                                                    & 16                                                     \\
\multirow{2}{*}{Xylo-Size}        & DoP          & 15                                                    & 15                                                     \\
                                  & Full         & 4                                                     & 4                                                      \\
\multirow{2}{*}{Full Spectrogram} & DoP          & 512                                                   & 512                                                    \\
                                  & Full         & 2048                                                  & 512                                                    \\ \bottomrule
\end{tabular}
\end{table}

We measure detection performance with respect to per-pixel accuracy, Area Under the Receiver Operating Curve (AUROC), Area Under the Precision Recall Curve (AUPRC) and F1-Score, focusing on the class-balanced metrics AUPRC and F1-Score in particular, in line with other literature \cite{mesarcik_learning_2022, pritchard_spiking_2024}.
\section{Computational Power Requirement Modelling}
A key hindrance to including machine-learning-based methods in operational radio astronomy observation is the computational requirements, which are significantly guided by energy usage.
We provide three energy usage estimates: the first is a lower-bound based on the spiking activity of the trained network; the second is an upper-bound based on the maximum power draw of Xylo hardware ($550\mu W$), and a final estimate combining the measured idle power-draw of the Xylo hardware $216\mu W$ \cite{bos_sub-mw_2022} and the lower-bound estimate.

We estimate a proxy-number of FLOPs per layer $l$, based on known literature\cite{barchid_spiking_2023} as:
\begin{equation}\label{eq:flop_model}
    FLOPs_{SNN}(l) = C_{in} \times C_{out} \times R_s(l)
\end{equation}
where $C_{in}$ is the number of channels in, $C_{out}$ is the number channels out and $R_s(l)$ is the spiking rate of layer $l$.
We measure the average spike rate of the entire network over the test dataset post-training to provide an indicative measurement. 
We calculate a lower-bound power usage estimate by multiplying $FLOPs_{SNN}$ by $0.9pJ$, the power usage of a single add operation at 40nm CMOS technology\cite{horowitz_11_2014}.

An upper estimate of power usage for an instrument with $N$ channels and $P$ polarisations on the Xylo hardware is given by:
\begin{equation}\label{eq:power_model}
    \frac{N \cdot P}{16} \cdot 550\,\mu W
\end{equation} for either no or full polarisation with the result for DoP polarisation given by:
\begin{equation}\label{eq:power_model_DOP}
    \lceil\frac{N}{15}\rceil \cdot 550\,\mu W.
\end{equation}

Finally, we provide a balanced power usage estimate by adding the lower-bound estimate to the idle power draw of the Xylo hardware.
These equations provide power-usage metrics for inference over a single spectrogram patch.
To provide estimates for an entire spectrogram, we multiply the full-polarisation values by $(\frac{512}{4})^2$ and the DoP metrics by $\lceil \frac{512}{15}\rceil^2$ to build an estimate for the energy required by a single chip to inference over an entire spectrogram.
By providing a lower, upper, and balanced estimate, we move towards characterising the power usage of neuromorphic techniques in radio astronomy data processing.
All code is available online\footnote{\url{https://github.com/pritchardn/SNN-RFI-SUPER/tree/polarization}}.
\section{Results}
Table \ref{tab:results} presents detection performance metrics, averaged over 10 trial runs for each model and polarisation configuration.
The results indicate that the Latency + Divisive Normalisation (DN) configuration, when patched to 16 channels, achieves the highest detection accuracy.
This suggests that leveraging latency encoding in conjunction with divisive normalisation enhances the model’s ability to discriminate between RFI and astronomical signals.
However, including the entire spectrogram as input hinders performance, likely due to the introduction of redundant or less informative spectral features without increasing the size of hidden layers, which may increase model complexity without contributing to meaningful signal differentiation.
It is likely that a larger model will achieve greater performance given more training time.
Additionally, divisive normalisation improves detection accuracy when full polarisation data is used but paradoxically leads to a decline in performance when applied in conjunction with the Degree of Polarisation (DoP) input.
This suggests that the normalisation process amplifies relevant polarisation features in one case while possibly suppressing useful signal variations in the other.
Lastly, truncating the model to fit within the constraints of the Xylo neuromorphic hardware results in minor performance degradation.
However, in the specific case of full polarisation with divisive normalisation, performance remains very close to the best-performing models, suggesting that carefully selected preprocessed strategies can mitigate the impact of hardware-imposed limitations.
\begin{table*}[!htbp]
\centering
\caption{Final results averaged over 10 trials (average and standard-deviation) for all model and polarisation configurations (Full or DoP) and with(out) Divisive Normalisation (DN). Best scores in bold. Second-best scores underlined.}
\label{tab:results}
\begin{tabular}{@{}cccccccccc@{}}
\toprule
Model Type        & Polarisation & \multicolumn{2}{c}{Accuracy} & \multicolumn{2}{c}{AUROC} & \multicolumn{2}{c}{AUPRC} & \multicolumn{2}{c}{F1} \\ \midrule
Latency           & DoP          & 0.974         & 0.004        & 0.942       & 0.021       & 0.793       & 0.034       & 0.766      & 0.034     \\
Latency + DN      & DoP          & 0.963         & 0.004        & 0.810       & 0.007       & 0.616       & 0.026       & 0.602      & 0.028     \\
Latency           & Full         & 0.978         & 0.010        & 0.933       & 0.011       & 0.771       & 0.071       & 0.725      & 0.102     \\
Latency + DN      & Full         & \textbf{0.997}         & 0.001        & \textbf{0.977}       & 0.002       & \textbf{0.960}       & 0.071       & \textbf{0.955}      & 0.031     \\
Latency-Xylo      & DoP          & 0.968         & 0.006        & 0.948       & 0.010       & 0.789       & 0.031       & 0.744      & 0.037     \\
Latency-Xylo + DN & DoP          & 0.939         & 0.010        & 0.791       & 0.008       & 0.524       & 0.032       & 0.474      & 0.044     \\
Latency-Xylo      & Full         & 0.943         & 0.052        & 0.932       & 0.029       & 0.694       & 0.113       & 0.546      & 0.197     \\
Latency-Xylo + DN & Full         & \underline{0.996}         & 0.000        & \underline{0.966}       & 0.003       & \underline{0.941}       & 0.113       & \underline{0.936}      & 0.007     \\
Latency-Full      & DoP          & 0.961         & 0.046        & 0.885       & 0.034       & 0.770       & 0.095       & 0.743      & 0.145     \\
Latency-Full + DN & DoP          & 0.982         & 0.001        & 0.906       & 0.009       & 0.831       & 0.014       & 0.826      & 0.014     \\
Latency-Full      & Full         & 0.963         & 0.025        & 0.807       & 0.022       & 0.628       & 0.075       & 0.609      & 0.102     \\
Latency-Full + DN & Full         & 0.994         & 0.001        & 0.964       & 0.005       & 0.933       & 0.075       & 0.931      & 0.031     \\ \bottomrule
\end{tabular}
\end{table*}

Table \ref{tab:results_comparison} presents a comparative analysis of our best-performing model, patched with divisive normalisation and full-polarisation, against state-of-the-art RFI detection schemes evaluated on a principally identical dataset.
The results demonstrate, for the first time, that a Spiking Neural Network (SNN) achieves state-of-the-art performance, surpassing traditional operational RFI detection methods and artificial neural network (ANN) approaches and ANN-to-SNN (ANN2SNN) conversion techniques.
This result highlights the potential of directly trained SNNs as a viable, high-performance alternative to conventional machine learning-based RFI detection, particularly in scenarios where low latency and energy efficiency are critical.
\begin{table}[!htbp]
\centering
\caption{Results compared to other SNN-based RFI detection methods. Divisive Normalisation abbreviated to `DN'. Best scores in bold.}
\label{tab:results_comparison}
\begin{tabular}{@{}ccccc@{}}
\toprule
Work      & Method            & AUROC & AUPRC & F1-Score \\ \midrule
\cite{mesarcik_learning_2022}      & AOFlagger          & 0.973 & 0.880 & 0.873    \\
\cite{mesarcik_learning_2022}      & Auto-Encoder          & 0.981 & 0.927 & 0.910    \\
\cite{pritchard_rfi_2024}      & Auto-Encoder          & 0.994 & 0.959 & 0.945    \\
\cite{pritchard_rfi_2024}      & ANN2SNN          & 0.944 & 0.910 & 0.953    \\
\cite{pritchard_supervised_2024}     & BPTT & 0.929 & 0.785 & 0.761    \\
\cite{pritchard_spiking_2024}  & BPTT + DN & 0.968 & 0.914 & 0.907    \\
This work & Full Polarisation               & \textbf{0.997}     & \textbf{0.960}     & \textbf{0.955}        \\ \bottomrule
\end{tabular}
\end{table}

We finally provide energy usage estimates for our Xylo-sized models in Table \ref{tab:energy_usage}.
Despite the increased flop count and per-patch energy usage of the DoP style model, the increased input width (in real spectrogram channels) compared to the full polarisation method results in a significantly lower total energy-usage estimate, accompanied by a drop in detection performance.
These results show that even under an upper-bound estimate, total power consumption is tiny compared to a small GPU-based system like an Nvidia Jetson Nano with a nominal power consumption of $\approx5W$\footnote{\url{https://docs.nvidia.com/jetson/archives/l4t-archived/l4t-3275/index.html}}.
Despite these results being estimates, the potential improvements are so enormous that neuromorphic computing needs to be considered seriously for high-performance scientific signal processing applications, including radio astronomy.
\begin{table*}[!htbp]
\centering
\caption{Energy usage estimates for Xylo-scale models performing RFI detection on a full HERA spectrogram.}
\label{tab:energy_usage}
\begin{tabular}{@{}cccccc@{}}
\toprule
Model Type                        & $FLOPs_{SNN}$ & Patch-Energy ($\mu W$) & Lower ($mW$) & Upper ($mW$) & Balanced ($mW$) \\\midrule
Xylo-Polarised          & 827392          & 1.4                                                   & 22.9 & 64 &  53.6                                                    \\
Xylo-DoP        & 1642496          &  2.79                                                   & 3.42 & 17.5 &  11.8                                                    \\ \bottomrule
\end{tabular}
\end{table*}
These preliminary results highlight the potential for neuromorphic hardware to deliver competitive RFI detection performance while maintaining low power requirements, making it a compelling option for deployment in power-constrained observatory environments.
\section{Conclusions}
This work demonstrates the integration of polarisation information into spiking neural networks (SNNs) for radio frequency interference (RFI) detection, marking an advancement in machine learning-based approaches for radio astronomy.
We evaluate performance on a known benchmark dataset and show that a supervised SNN can achieve state-of-the-art detection performance, surpassing previous methods.
Even when constrained to the computational limitations of SynSense's Xylo hardware, our models still perform comparably to the previous best SNN-based RFI detection methods, reinforcing the viability of neuromorphic computing for real-time astronomical signal processing.

These models not only achieve state-of-the-art detection accuracy but do so with significantly reduced computational requirements.
Our estimates suggest they operate using only 10s of mW per spectrogram, representing improvements by orders of magnitude, rather than incremental refinements.
Given the multi-decadal lifespan of modern radio observatories, such efficiency improvements could lead to substantial reductions in operational costs and power consumption, making neuromorphic approaches worthwhile exploring.

Future work will focus on investigating more sophisticated network architectures, applying SNNs to broader data processing challenges in radio astronomy, and benchmarking on real neuromorphic hardware to validate power and performance estimates this work presents.
\section*{Acknowledgements}
This work was supported by a Westpac Future Leaders Scholarship, an Australian Government Research Training Program Fees Offset and an Australian Government Research Training Program Stipend.
This work was supported by resources provided by the Pawsey Supercomputing Centre with funding from the Australian Government and the Government of Western Australia. 
\FloatBarrier
\printbibliography
\end{document}